\documentclass[10pt,conference]{IEEEtran}

\usepackage{cite}
\usepackage{amsmath,amssymb,amsfonts}
\usepackage{graphicx}
\usepackage{textcomp}
\usepackage{xcolor}
\usepackage{booktabs}
\usepackage{array}
\usepackage{multirow}
\usepackage{placeins}
\usepackage{microtype}
\usepackage[hidelinks]{hyperref}
\hypersetup{pdfauthor={Haobin Chen, Ao Chang, Heqin Zhu, Rundong Wang, Ting Liu, Shaohua Kevin Zhou}}
\usepackage[capitalise,nameinlink]{cleveref}

\def\BibTeX{{\rm B\kern-.05em{\sc i\kern-.025em b}\kern-.08em
    T\kern-.1667em\lower.7ex\hbox{E}\kern-.125emX}}
\newcommand{\ours}{MARR}
\renewcommand{\IEEEtitletopspaceextra}{-1.0\baselineskip}

\begin{document}

\title{\ours: Decoupling Policy, Execution, and Calibration for All-in-One Medical Image Restoration}

\author{
\IEEEauthorblockN{Haobin Chen\textsuperscript{1,2}, Ao Chang\textsuperscript{1,2}, Heqin Zhu\textsuperscript{1,2}, Rundong Wang\textsuperscript{1,2}, Ting Liu\textsuperscript{1,2}, Shaohua Kevin Zhou\textsuperscript{1,2,3,4,5,*}}
\IEEEauthorblockA{\scriptsize
\textsuperscript{1}School of Biomedical Engineering, Division of Life Sciences and Medicine, University of Science and Technology of China (USTC), Hefei, Anhui, 230026, China\\
\textsuperscript{2}Medical Imaging, Robotics, Analytic Computing \& Learning (MIRACLE) Lab, YRD-RIGHT, Suzhou Institute for Advanced Research, USTC, Suzhou, Jiangsu, 215123, China\\
\textsuperscript{3}Biomedical Basic Research Center (BBRC) of Jiangsu, Suzhou, Jiangsu, 215123, China\\
\textsuperscript{4}Jiangsu Key Laboratory of Multimodal Digital Twin Technology, Suzhou, Jiangsu, 215123, China\\
\textsuperscript{5}State Key Laboratory of Precision and Intelligent Chemistry, Hefei, Anhui, 230026, China\\
Emails: \{aleck, aochang, wangrundong, ting\_liu\}@mail.ustc.edu.cn; zhuheqin@ustc.edu; s.kevin.zhou@gmail.com \quad (\textsuperscript{*}Corresponding author)}
}

\maketitle

\begin{abstract}
All-in-one medical image restoration seeks to recover heterogeneous clinical images with a single model, but PET, CT, and MRI differ substantially in degradation statistics, anatomical contrast, and output-space bias. A fully shared network can entangle modality-specific residual errors, whereas separate modality-specific networks sacrifice the practical advantages of unified deployment. We therefore recast all-in-one restoration as a question of where limited adaptation should be placed: policy selection, feature execution, or output calibration. We propose MARR, a compact restoration framework that constrains multi-modality adaptation into degradation-aware policy routing, modality-private residual execution, and image-domain residual correction without requiring degradation labels or separate modality-specific models. The policy branch forms a routing prompt from input statistics, latent content, and modality identity, and uses it only as a control signal. Prompt-gated modality-private adapters then perform lightweight residual refinement at intermediate decoder stages, while zero-initialized modality-specific output heads calibrate the final image-domain residual without perturbing the initial shared prediction. On an all-in-one PET, CT, and MRI restoration benchmark, MARR outperforms thirteen methods re-trained under the same protocol, achieving PSNR values of 37.34 dB, 33.85 dB, and 32.09 dB on PET, CT, and MRI, respectively, and the best modality-average PSNR of 34.43 dB.The code is publicly available at \url{https://github.com/CHB-learner/MARR}.
\end{abstract}

\begin{IEEEkeywords}
all-in-one medical image restoration, PET, CT, MRI, prompt learning, modality adaptation, residual calibration
\end{IEEEkeywords}

\section{Introduction}
Medical image restoration is a core preprocessing step for downstream diagnosis, quantitative analysis, and image-guided intervention. In clinical pipelines, however, each imaging modality presents a distinct restoration failure mode. PET restoration must suppress low-count noise without erasing weak functional uptake, low-dose CT denoising must remove structured artifacts while preserving calibrated intensity, and MRI super-resolution must recover anatomical edges and tissue contrast. Training a separate model for each modality is technically feasible, but it increases validation cost, fragments maintenance, and complicates deployment in multi-modality clinical environments.

All-in-one restoration addresses this deployment problem by training one model across multiple degradations and modalities. Recent image restoration networks have demonstrated strong generic capacity through transformers, shifted-window attention, nonlinear activation-free blocks, and prompt learning \cite{Zamir2021RestormerEfficientTransformer,Liang2021SwinIRImageRestoration,chen2022simple,potlapalli2023promptir}. Medical all-in-one methods extend this idea with task routing, task-adaptive attention, and degradation-aware prompts for PET, CT, MRI, and related restoration tasks \cite{Wei2025DegradationAwarePromptedTransformer,Yang2025TATTaskAdaptiveTransformer,Yang2024AllInOneMedicalImage,Yang2024RegionAttentionTransformer}. These designs show that task and degradation cues are valuable, but they often leave the functional role of the cue implicit. In particular, the same prompt may be expected simultaneously to select a restoration behavior, modulate intermediate features, and correct the final reconstruction.

Rather than adding another unconstrained conditioning module, we revisit all-in-one medical restoration as a practical design problem: where should limited adaptation occur in a shared model? A unified network can benefit from first selecting a degradation-aware policy, then executing modality-specific refinement without overwriting shared anatomical priors, and finally correcting residual image-domain bias. This separation is important because PET, CT, and MRI differ not only in degradation strength, but also in intensity scale, texture distribution, and the residual errors that remain after shared restoration. Treating these roles separately makes the model easier to audit and reduces the risk that a single prompt branch absorbs incompatible responsibilities.

To this end, we propose \ours, a modality-aware residual recalibration framework for practical all-in-one medical image restoration. As shown in Fig.~\ref{fig:overview}, \ours{} contains three coordinated components. First, a degradation-aware prompt policy estimator converts image statistics, modality metadata, and latent content into a policy prompt by routing over modality and degradation prompt pools. Second, prompt-gated modality-private adapters use this policy prompt to inject conservative residual refinements at decoder levels where modality-specific details are reconstructed. Third, zero-initialized modality-specific output heads perform final residual calibration in image space. The resulting system keeps common restoration priors shared while assigning explicit and lightweight roles to decision, execution, and correction.

Our contributions are summarized as follows.
\begin{itemize}
    \item We formulate unified PET, CT, and MRI restoration as a shared-private adaptation problem shaped by modality-dependent degradation, anatomical contrast variation, and feature interference.
    \item We propose a compact and deployable framework that decomposes all-in-one restoration into adaptive policy routing, modality-aware residual execution, and output-space calibration.
    \item We design a lightweight routing-and-adaptation mechanism that uses image statistics, latent content, and modality identity to guide modality-aware restoration while preserving shared anatomical priors.
    \item We demonstrate consistent gains over thirteen representative methods on PET synthesis, CT denoising, and MRI super-resolution under a unified evaluation protocol.
\end{itemize}

\FloatBarrier
\section{Related Work}
\subsection{Generic Image Restoration}
Transformer-based restoration networks build on the broader success of attention and vision transformers \cite{Vaswani2017AttentionisAll,Dosovitskiy2020AnImageis,Liu2021SwinTransformerHierarchical}. SwinIR \cite{Liang2021SwinIRImageRestoration} and Restormer \cite{Zamir2021RestormerEfficientTransformer} established strong baselines for image denoising, super-resolution, and deblurring by combining long-range modeling with efficient local computation. NAFNet \cite{chen2022simple} showed that a carefully designed convolutional architecture without explicit nonlinear activations can also provide a competitive restoration backbone. More recent state-space and recurrent-style visual models have explored efficient long-range modeling for restoration and medical imaging, including Mamba \cite{Gu2023MambaLinearTimeSequence}, Vision Mamba \cite{Zhu2024VisionMambaEfficient}, VMamba \cite{Liu2024VMambaVisualState}, MambaIR \cite{Guo2024MambaIRASimple}, MambaIRv2 \cite{Guo2024MambaIRv2AttentiveState}, and Restore-RWKV \cite{Yang2024RestoreRWKVEfficientand}. PromptIR \cite{potlapalli2023promptir} further introduced prompt learning for all-in-one blind restoration, demonstrating that learned prompt representations can help a single model adapt to multiple degradations. These methods motivate shared restoration backbones, but most are designed for natural images and do not explicitly separate modality routing, modality-specific feature specialization, and output calibration.

\subsection{All-in-One Medical Image Restoration}
Medical image restoration has developed along modality-specific lines. PET studies address synthesis and denoising with adversarial, segmentation-guided, multi-center, and diffusion-based models \cite{Chan2018NoiseAdaptiveDeep,Luo2021Adaptiverectificationbased,Zhou20223DSegmentationGuided,Yang2023DRMCAGeneralist,Jing2026MAPDiffMultiAnchorGuided}. CT and MRI restoration have also been studied through transformer encoder-decoder denoisers, deep unfolding, and hybrid-domain restoration networks \cite{Guo2024TransformerEncoderDecoderwith,Wang2025DualDomainSelfConsistencyEnhancedDeep,Liu2026HDFNetHybriddomainfusionnetwork}. Beyond single-modality restoration, medical all-in-one restoration has recently attracted increasing attention. AMIR proposes task-adaptive routing for unified medical restoration \cite{Yang2024AllInOneMedicalImage}, RAT studies region attention for medical image restoration \cite{Yang2024RegionAttentionTransformer}, and TAT introduces task-adaptive transformer modeling for all-in-one restoration \cite{Yang2025TATTaskAdaptiveTransformer}. DaPT uses degradation-aware prompts for unified medical image restoration \cite{Wei2025DegradationAwarePromptedTransformer}, and VQ-codebook and diffusion-based methods explore stronger priors for all-in-one restoration and synthesis \cite{Chen2025AllinOneMedicalImage,Tang2026FastandSlow,Wei2025RethinkingDiffusionBridge}. These studies show that task and domain cues are crucial. Our work focuses on a complementary practical design: separating policy selection, modality-private residual execution, and output residual calibration within one deployable restoration model.

\subsection{Prompting and Lightweight Adaptation}
Prompt learning offers a compact mechanism for encoding degradation or task information in restoration networks. Recent restoration works use prompts, focal modulation, LoRA-inspired attention, or quality-conditioned pseudo-labeling to improve adaptation under diverse degradations \cite{Cui2025FocalModulationfor,Cui2026CDIRLoRAInspiredAttention,Xiao2026QualiTeacherQualityConditionedPseudoLabeling}. Adapter-style modules provide another way to specialize a shared model while keeping most parameters shared. We do not treat prompt routing, adapters, or residual heads as isolated primitives. The distinction of \ours{} is to constrain their functions: the prompt branch supplies control, private adapters absorb feature-space residuals, and output heads correct image-domain bias. This role separation turns otherwise independent add-ons into an auditable shared-private restoration mechanism.

\FloatBarrier
\section{Method}
\subsection{Problem Formulation and Overview}
Let $x$ denote a degraded medical image, $m \in \{\mathrm{PET}, \mathrm{CT}, \mathrm{MRI}\}$ denote its modality, and $y$ denote the clean target. The goal is to learn a unified restoration function
\begin{equation}
    \hat{y} = f_{\theta}(x,m)
\end{equation}
that restores all modalities using one model. The key challenge is to preserve shared low-level priors while avoiding modality interference.

\begin{figure*}[t]
\centering
\includegraphics[width=0.95\textwidth]{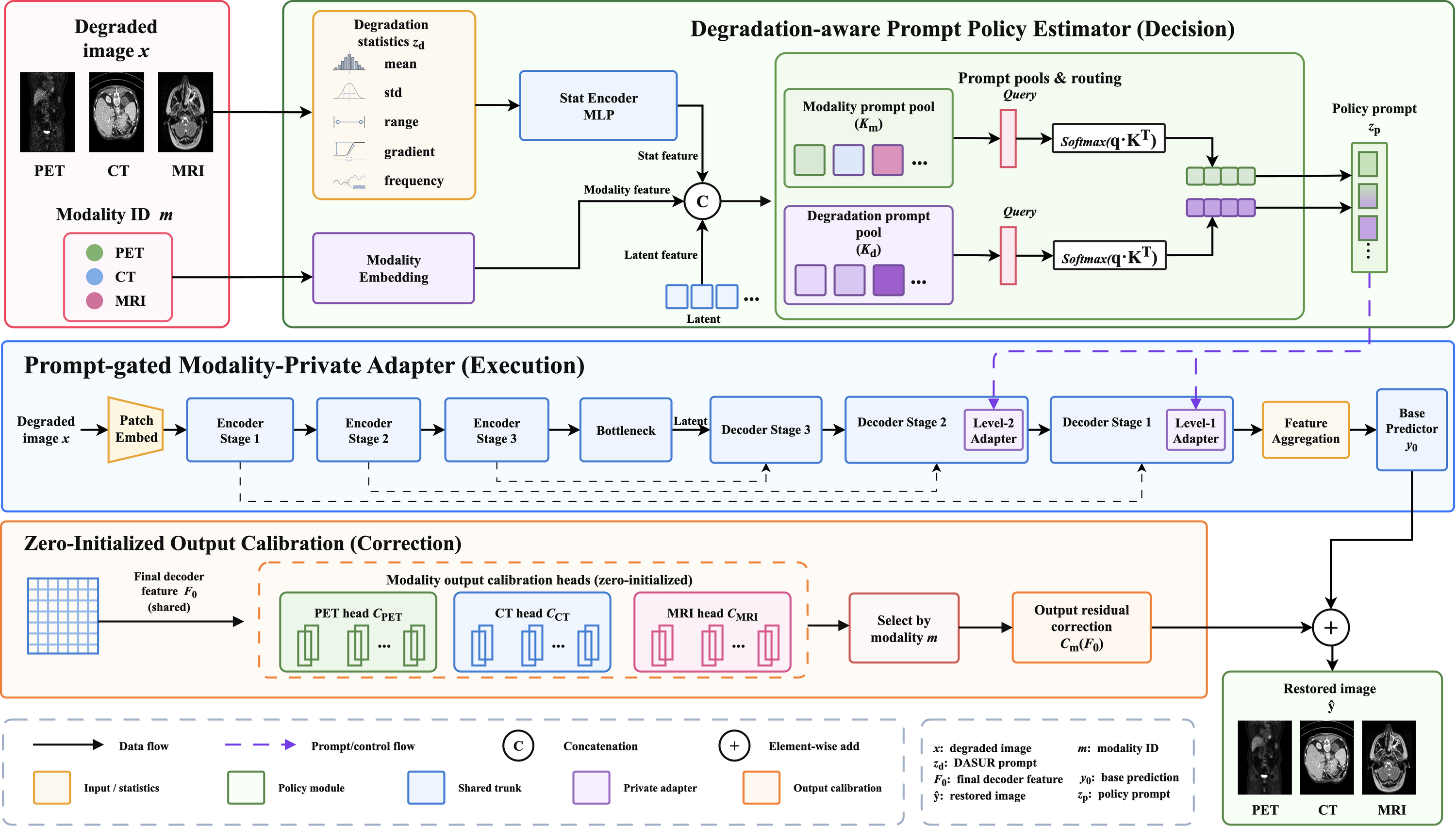}
\caption{Overall architecture of \ours{}. The framework separates all-in-one restoration into three roles: degradation-aware prompt policy estimation for decision making, prompt-gated modality-private adaptation for execution, and zero-initialized output calibration for correction. Solid arrows denote restoration data flow and dashed arrows denote prompt/control flow.}
\label{fig:overview}
\end{figure*}

\ours{} implements this unified model with three restricted adaptation paths in the MA-DASUR network. The policy path estimates a degradation-aware prompt $z_p$ that controls restoration behavior. The execution path applies prompt-gated modality-private residual adapters to specialize intermediate decoder features. The correction path uses zero-initialized modality-specific output heads to correct final residual bias. Fig.~\ref{fig:overview} gives an overview of this decision-execution-correction decomposition.

\subsection{Design Rationale}
The central design choice of \ours{} is to separate the source of adaptivity from the place where adaptation is applied. In many all-in-one restoration models, a task embedding or prompt is inserted into the network as a generic condition. This design is compact, but it does not specify whether the condition selects a restoration behavior, specializes intermediate features, or corrects final image-domain bias. \ours{} makes these roles explicit in a deployable single-model design. The policy prompt supports decision making, the private adapters perform modality-specific execution, and the output heads provide residual calibration.

This separation is implemented as a role constraint rather than as an unconstrained module stack. The policy branch does not inject image features or synthesize image details; it only produces control vectors. The private adapters can modify decoder features, but only through small residual perturbations. The output heads can correct the final image, but only through zero-initialized residual calibration. These restrictions make the decomposition auditable because each path is assigned a limited error type, reducing the risk of conflating task selection, feature specialization, and image-domain bias correction.

This decomposition is motivated by the complementary failure modes of PET, CT, and MRI. PET restoration is highly sensitive to noise statistics and weak functional structures; CT restoration is sensitive to small intensity bias and local artifact suppression; and MRI restoration requires preserving anatomical texture under large sample imbalance. A single shared feature transformation is unlikely to handle these residual errors uniformly. At the same time, a fully separate model per modality would discard shared anatomical and low-level restoration priors. \ours{} therefore uses one shared restoration path for common cues and adds lightweight private capacity only where modality-specific residuals are most likely to appear.

\subsection{Degradation-Aware Prompt Policy}
The prompt policy estimator converts three complementary cues into a compact routing prompt: degradation statistics from the input image, modality identity from metadata, and latent content from the bottleneck feature. For each input image, we first compute the statistic vector
\begin{equation}
    s(x)= [\mu(x), \sigma(x), r(x), e_g(x), \rho_f(x)],
\end{equation}
where $\mu$ and $\sigma$ are the image mean and standard deviation, $r$ is the intensity range, $e_g$ is the average gradient energy, and $\rho_f$ is the high-to-low frequency energy ratio. These statistics capture intensity dispersion, local edge strength, and frequency-domain degradation tendency.

The statistic vector is deliberately simple. It does not require an external degradation label or a separate quality estimator, which makes it suitable for all-in-one restoration where degradation type and strength may vary across modalities. In Fig.~\ref{fig:overview}, this branch is denoted as the degradation statistic descriptor $z_d=\phi_s(s(x))$. It is an input-derived statistic feature rather than another modality label. In particular, $e_g$ emphasizes local structural corruption, while $\rho_f$ distinguishes high-frequency noise or texture loss from low-frequency intensity drift. These cues complement modality identity because two samples from the same modality can still have different degradation severity.

The policy network also extracts a latent content descriptor $c$ from the bottleneck feature using global pooling and obtains a learnable modality embedding $e_m$. These three factors form a query
\begin{equation}
    q = \phi_q([e_m, z_d, \phi_c(c)]),
\end{equation}
where $\phi_s$, $\phi_c$, and $\phi_q$ are lightweight multilayer projections. The query selects prompts from a modality prompt pool $P_m$ and a degradation prompt pool $P_d$:
\begin{equation}
    a_m = \operatorname{softmax}\left(\frac{P_m q}{\sqrt{d}}\right), \quad
    a_d = \operatorname{softmax}\left(\frac{P_d q}{\sqrt{d}}\right).
\end{equation}
The selected modality prompt and degradation prompt are then fused with the query context to form the policy prompt $z_p$. The routing operation is a soft selection over learned prompt pools. Its purpose is not to introduce a new attention operator, but to constrain the routed prompt to act as a control signal for residual adapters and calibration rather than as a generic feature token or image-feature injector.

To avoid prompt collapse, we use a lightweight policy regularizer that encourages prompt diversity and discourages prematurely peaked routing distributions:
\begin{equation}
    \mathcal{L}_{p} = \mathcal{L}_{div} + \lambda_e \mathcal{L}_{ent}.
\end{equation}
The diversity term discourages prompt vectors from becoming redundant, and the entropy term prevents the routing distribution from degenerating too early in training. This regularization is not intended to dominate the reconstruction objective. It serves as a small stabilizer so that the prompt pool remains a meaningful decision space.

\subsection{Prompt-Gated Modality-Private Residual Adaptation}
\begin{figure}[t]
\centering
\includegraphics[width=0.60\columnwidth]{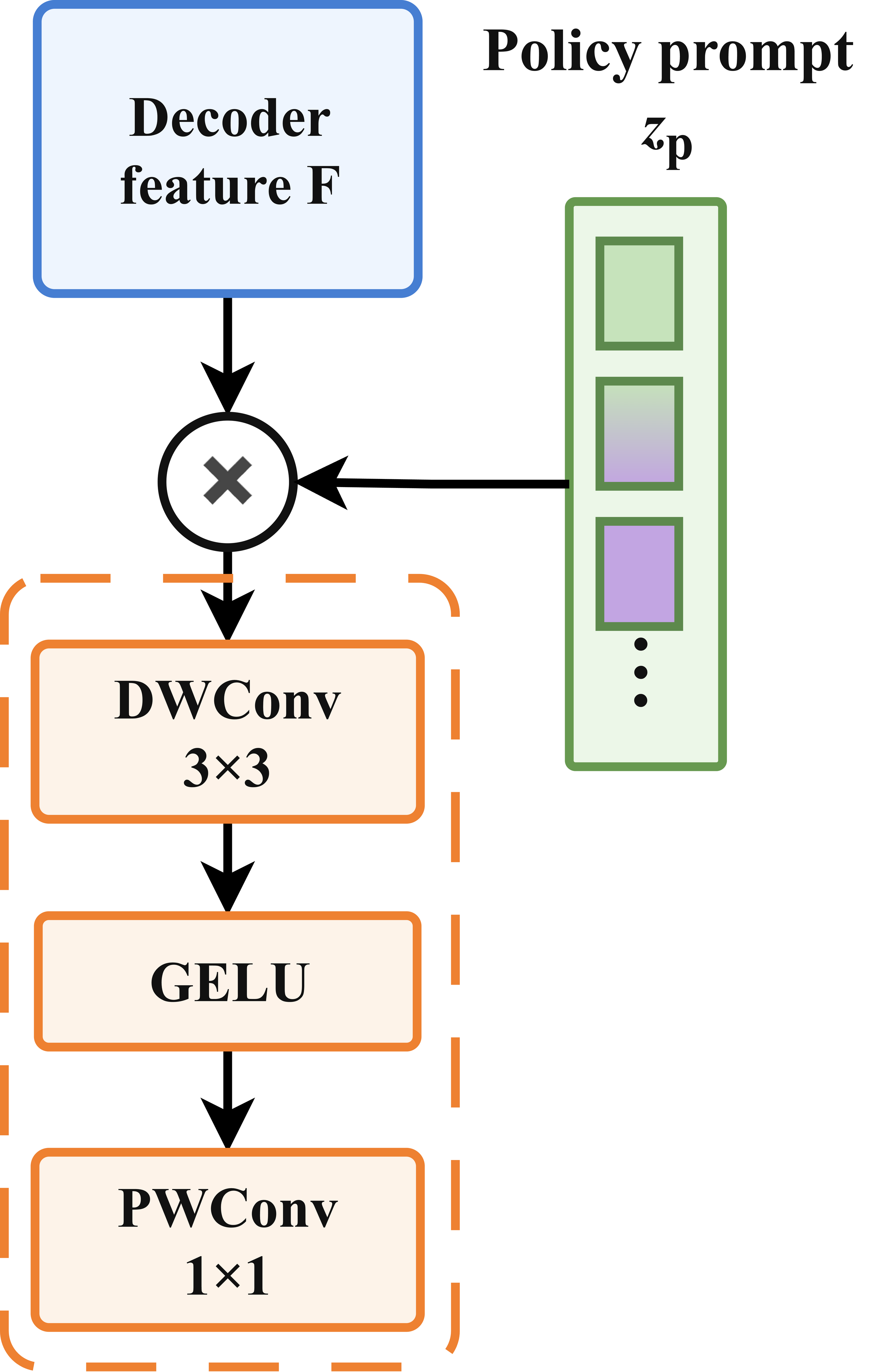}
\caption{Prompt-gated modality-private adapter. The policy prompt $z_p$ predicts a modality gate that controls a lightweight DWConv-GELU-PWConv residual branch for the current decoder feature $F$.}
\label{fig:adapter}
\end{figure}

The policy prompt controls modality-private adapters at intermediate decoder stages. Fig.~\ref{fig:adapter} shows the adapter used at each selected decoder level. For a decoder feature $F_l$ at level $l$, \ours{} applies a modality-specific residual adapter $A_m^l(\cdot)$ and a prompt-dependent gate $g_m^l(z_p)$:
\begin{equation}
    \tilde{F}_l = F_l + \alpha \, g_m^l(z_p) \, A_m^l(F_l).
\end{equation}
Each adapter is implemented as a lightweight depthwise-pointwise residual block, and the prompt gate is generated from $z_p$ by a small multilayer projection. Concretely, each modality owns one DWConv-GELU-PWConv residual branch, while the policy prompt predicts modality gates through a sigmoid projection. The final pointwise convolution is initialized to zero, and the residual scale $\alpha$ is kept small, so the branch starts as an identity-preserving perturbation rather than an uncontrolled modality-specific transformation. We place adapters at two decoder stages, giving the network modality-private capacity where semantic and structural details are reconstructed while keeping the main representation shared.

This module provides the execution component of \ours{}. The shared trunk learns modality-agnostic restoration priors, while the private adapters absorb modality-specific residual corrections. The residual form also makes adaptation conservative and lightweight. If a modality does not need additional correction at a stage, the gated residual can remain small.

Adapter placement is also important. Placing private modules too early can overfit modality-specific intensity statistics before shared texture and structure are formed, while placing them only at the output can be insufficient for recovering modality-specific details. We therefore insert adapters in intermediate decoder stages, where the representation already contains bottleneck context but still retains enough spatial resolution for local refinement. The prompt gate allows the same modality to receive different residual strengths under different degradation statistics.

\subsection{Zero-Initialized Output Calibration}
Even after feature-level adaptation, the final reconstruction can retain modality-dependent residual bias. We therefore introduce modality-specific output calibration heads. Given the final decoder feature $F_o$ and the base prediction $\hat{y}_0$, the final output is
\begin{equation}
    \hat{y} = \hat{y}_0 + \beta C_m(F_o),
\end{equation}
where $C_m$ is the output head for modality $m$. All output heads are initialized with zero weights and zero bias. At initialization, the model prediction is exactly $\hat{y}_0$, and the calibration branch learns only residual corrections during training. This design avoids early training disturbance while allowing the model to correct PET, CT, and MRI output-space bias separately.

The zero-initialized design is useful for stable optimization. Without this constraint, a modality-specific output head can immediately perturb the image intensity distribution before the shared restoration representation is well trained. By starting from an identity residual path, the calibration branch behaves as a late-stage bias corrector. It can learn small modality-specific corrections for CT intensity bias, PET noise residuals, and MRI texture errors, while preserving the common restoration signal produced by the shared trunk. Because this branch operates only at the output, it is not intended to replace feature-level adaptation; instead, it handles residual intensity or texture bias that remains after the shared prediction and private adapters have formed the main anatomical structure.

\subsection{Training Objective}
We train \ours{} with a modality-balanced reconstruction objective and the policy regularizer:
\begin{equation}
    \mathcal{L} = \sum_{m} w_m \, \mathbb{E}_{(x,y)\in \mathcal{D}_m}
    \left[\|\hat{y}-y\|_1\right] + \lambda_p \mathcal{L}_p.
\end{equation}
The modality weights compensate for the imbalance between PET, CT, and MRI samples. In our implementation, the policy regularization weight is small, so the reconstruction objective remains dominant while prompt routing remains stable.

\begin{table*}[t]
\centering
\caption{All-in-One medical image restoration results. Avg. denotes the arithmetic mean over PET, CT, and MRI.}
\label{tab:main_results}
\setlength{\tabcolsep}{2.6pt}
\scriptsize
\resizebox{\textwidth}{!}{%
\begin{tabular}{lcccccccccccc}
\toprule
\multirow{2}{*}{Method} &
\multicolumn{3}{c}{PET Synthesis} &
\multicolumn{3}{c}{CT Denoising} &
\multicolumn{3}{c}{MRI Super-Resolution} &
\multicolumn{3}{c}{Avg.} \\
\cmidrule(lr){2-4}\cmidrule(lr){5-7}\cmidrule(lr){8-10}\cmidrule(lr){11-13}
& PSNR$\uparrow$ & SSIM$\uparrow$ & RMSE$\downarrow$
& PSNR$\uparrow$ & SSIM$\uparrow$ & RMSE$\downarrow$
& PSNR$\uparrow$ & SSIM$\uparrow$ & RMSE$\downarrow$
& PSNR$\uparrow$ & SSIM$\uparrow$ & RMSE$\downarrow$ \\
\midrule
NAFNet & 35.15\textsuperscript{*} & 0.9259\textsuperscript{*} & 0.1105\textsuperscript{*} & 33.39\textsuperscript{*} & 0.9164\textsuperscript{*} & 8.7455\textsuperscript{*} & 29.83\textsuperscript{*} & 0.9098\textsuperscript{*} & 37.3700\textsuperscript{*} & 32.79\textsuperscript{*} & 0.9173\textsuperscript{*} & 15.4086\textsuperscript{*} \\
Restormer & 35.36\textsuperscript{*} & 0.9276\textsuperscript{*} & 0.1080\textsuperscript{*} & 33.54\textsuperscript{*} & 0.9177\textsuperscript{*} & 8.5975\textsuperscript{*} & 30.09\textsuperscript{*} & 0.9136\textsuperscript{*} & 36.3138\textsuperscript{*} & 33.00\textsuperscript{*} & 0.9197\textsuperscript{*} & 15.0064\textsuperscript{*} \\
SwinIR & 35.89\textsuperscript{*} & 0.9318\textsuperscript{*} & 0.1009\textsuperscript{*} & 33.47\textsuperscript{*} & 0.9161\textsuperscript{*} & 8.6713\textsuperscript{*} & 29.85\textsuperscript{*} & 0.9107\textsuperscript{*} & 37.1287\textsuperscript{*} & 33.07\textsuperscript{*} & 0.9195\textsuperscript{*} & 15.3003\textsuperscript{*} \\
DaPT & 35.53\textsuperscript{*} & 0.9288\textsuperscript{*} & 0.1051\textsuperscript{*} & 33.45\textsuperscript{*} & 0.9161\textsuperscript{*} & 8.6889\textsuperscript{*} & 30.08\textsuperscript{*} & 0.9134\textsuperscript{*} & 36.3602\textsuperscript{*} & 33.02\textsuperscript{*} & 0.9194\textsuperscript{*} & 15.0514\textsuperscript{*} \\
RAT & 35.46\textsuperscript{*} & 0.9284\textsuperscript{*} & 0.1071\textsuperscript{*} & 33.52\textsuperscript{*} & 0.9174\textsuperscript{*} & 8.6280\textsuperscript{*} & 29.92\textsuperscript{*} & 0.9129\textsuperscript{*} & 36.9398\textsuperscript{*} & 32.96\textsuperscript{*} & 0.9196\textsuperscript{*} & 15.2250\textsuperscript{*} \\
ARGAN & 36.75\textsuperscript{*} & 0.9389\textsuperscript{*} & 0.0907\textsuperscript{*} & 32.92\textsuperscript{*} & 0.9111\textsuperscript{*} & 9.2110\textsuperscript{*} & 30.08\textsuperscript{*} & 0.9083\textsuperscript{*} & 35.7999\textsuperscript{*} & 33.25\textsuperscript{*} & 0.9194\textsuperscript{*} & 15.0339\textsuperscript{*} \\
DenoMamba & 36.81\textsuperscript{*} & 0.9367\textsuperscript{*} & 0.0895\textsuperscript{*} & 33.18\textsuperscript{*} & 0.9115\textsuperscript{*} & 8.9512\textsuperscript{*} & 30.32\textsuperscript{*} & 0.9091\textsuperscript{*} & 34.6972\textsuperscript{*} & 33.44\textsuperscript{*} & 0.9191\textsuperscript{*} & 14.5793\textsuperscript{*} \\
PromptIR & 37.25\textsuperscript{*} & 0.9477\textsuperscript{*} & 0.0859\textsuperscript{*} & 33.66\textsuperscript{*} & 0.9182\textsuperscript{*} & 8.4890\textsuperscript{*} & 31.87\textsuperscript{*} & 0.9381\textsuperscript{*} & 29.5561\textsuperscript{*} & 34.26\textsuperscript{*} & 0.9347\textsuperscript{*} & 12.7103\textsuperscript{*} \\
MambaIR & 37.17\textsuperscript{*} & 0.9458\textsuperscript{*} & 0.0864\textsuperscript{*} & 33.50\textsuperscript{*} & 0.9165\textsuperscript{*} & 8.6345\textsuperscript{*} & 31.31\textsuperscript{*} & 0.9305\textsuperscript{*} & 31.3150\textsuperscript{*} & 33.99\textsuperscript{*} & 0.9309\textsuperscript{*} & 13.3453\textsuperscript{*} \\
AirNet & 37.17\textsuperscript{*} & 0.9451\textsuperscript{*} & 0.0864\textsuperscript{*} & 33.62\textsuperscript{*} & 0.9176\textsuperscript{*} & 8.5226\textsuperscript{*} & 31.39\textsuperscript{*} & 0.9316\textsuperscript{*} & 31.1141\textsuperscript{*} & 34.06\textsuperscript{*} & 0.9314\textsuperscript{*} & 13.2410\textsuperscript{*} \\
AMIR & 37.12\textsuperscript{*} & 0.9475\textsuperscript{*} & 0.0876\textsuperscript{*} & 33.70\textsuperscript{*} & 0.9182\textsuperscript{*} & 8.4520\textsuperscript{*} & 32.03\textsuperscript{*} & 0.9396\textsuperscript{*} & 29.0988\textsuperscript{*} & 34.28\textsuperscript{*} & 0.9351\textsuperscript{*} & 12.5461\textsuperscript{*} \\
DATPRL-IR & 35.41\textsuperscript{*} & 0.9296\textsuperscript{*} & 0.1069\textsuperscript{*} & 33.43\textsuperscript{*} & 0.9154\textsuperscript{*} & 8.7084\textsuperscript{*} & 29.94\textsuperscript{*} & 0.9100\textsuperscript{*} & 36.8734\textsuperscript{*} & 32.93\textsuperscript{*} & 0.9183\textsuperscript{*} & 15.2296\textsuperscript{*} \\
TAT & 37.25\textsuperscript{*} & 0.9480\textsuperscript{*} & 0.0859\textsuperscript{*} & 33.79\textsuperscript{*} & 0.9192\textsuperscript{*} & 8.3642\textsuperscript{*} & 32.05\textsuperscript{*} & 0.9398\textsuperscript{*} & 29.0320\textsuperscript{*} & 34.36\textsuperscript{*} & 0.9357\textsuperscript{*} & 12.4940\textsuperscript{*} \\
\midrule
\ours{} & \textbf{37.34} & \textbf{0.9485} & \textbf{0.0849} & \textbf{33.85} & \textbf{0.9199} & \textbf{8.3469} & \textbf{32.09} & \textbf{0.9399} & \textbf{29.0116} & \textbf{34.43} & \textbf{0.9361} & \textbf{12.4811} \\
\bottomrule
\end{tabular}%
}
\vspace{1pt}
\parbox{\textwidth}{\scriptsize \textsuperscript{*} Denotes results that differ significantly from \ours{} under a paired $t$-test at $p < 0.05$.}
\end{table*}

Model selection uses a validation score that accounts for all modalities rather than only the sample-weighted average. This choice is important because the test set is dominated by MRI slices. A purely sample-weighted validation criterion can hide PET or CT degradation, whereas a modality-aware criterion better matches the all-in-one objective of improving every clinical modality.

\begin{table*}[t]
\centering
\caption{Ablation study results of \ours{}. Bold indicates the best result in each metric column.}
\label{tab:ablation}
\setlength{\tabcolsep}{2.7pt}
\scriptsize
\resizebox{\textwidth}{!}{%
\begin{tabular}{llcccccccccc}
\toprule
\multirow{2}{*}{Group} & \multirow{2}{*}{Variant} & \multirow{2}{*}{Params (M)} &
\multicolumn{3}{c}{PET Synthesis} &
\multicolumn{3}{c}{CT Denoising} &
\multicolumn{3}{c}{MRI Super-Resolution} \\
\cmidrule(lr){4-6}\cmidrule(lr){7-9}\cmidrule(lr){10-12}
& & & PSNR$\uparrow$ & SSIM$\uparrow$ & RMSE$\downarrow$
& PSNR$\uparrow$ & SSIM$\uparrow$ & RMSE$\downarrow$
& PSNR$\uparrow$ & SSIM$\uparrow$ & RMSE$\downarrow$ \\
\midrule
\multirow{3}{*}{Core modules}
& Output calibration only & 39.67 & 37.2644 & 0.9482 & 0.0857 & 33.7453 & 0.9185 & 8.4099 & 32.0037 & 0.9392 & 29.1782 \\
& Private adapter only & 39.75 & 37.2558 & 0.9481 & 0.0858 & 33.7476 & 0.9185 & 8.4080 & 32.0156 & 0.9394 & 29.1335 \\
& \ours{} & 39.75 & \textbf{37.3400} & \textbf{0.9485} & \textbf{0.0849} & \textbf{33.8500} & \textbf{0.9199} & \textbf{8.3469} & \textbf{32.0900} & \textbf{0.9399} & \textbf{29.0116} \\
\midrule
\multirow{5}{*}{Extended designs}
& + frequency adapter & 39.81 & 37.2405 & 0.9481 & 0.0861 & 33.7436 & 0.9184 & 8.4114 & 32.0075 & 0.9393 & 29.1621 \\
& + frequency prompt pool & 39.70 & 37.2498 & 0.9481 & 0.0859 & 33.7406 & 0.9185 & 8.4140 & 32.0167 & 0.9394 & 29.1392 \\
& + CT/MRI detail loss & 39.75 & 37.2844 & 0.9483 & 0.0856 & 33.7398 & 0.9184 & 8.4140 & 31.9888 & 0.9390 & 29.2256 \\
& + CT/MRI anchor & 39.75 & 37.2484 & 0.9481 & 0.0859 & 33.7414 & 0.9184 & 8.4138 & 32.0294 & 0.9396 & 29.0938 \\
& Full extended MFR & 39.85 & 37.2311 & 0.9480 & 0.0861 & 33.7429 & 0.9184 & 8.4126 & 32.0279 & 0.9396 & 29.1041 \\
\bottomrule
\end{tabular}%
}
\end{table*}

\section{Experiments}
\subsection{Dataset and Evaluation Metrics}
We evaluate on an all-in-one PET, CT, and MRI restoration benchmark organized as paired low-quality and high-quality slices. The train/validation/test splits contain 27,837/684/2,044 PET pairs, 18,351/128/211 CT pairs, and 40,500/5,800/11,400 MRI pairs. Training files are stored as binary slices, while validation and test files are stored as NIfTI slices with identical LQ/HQ filenames. Following common medical restoration practice, we report peak signal-to-noise ratio (PSNR), structural similarity (SSIM), and root mean squared error (RMSE). Higher PSNR and SSIM values are better, whereas lower RMSE values are better.

Because the test set is modality-imbalanced, Table~\ref{tab:main_results} reports the arithmetic average over PET, CT, and MRI. This average reflects modality-balanced performance across the three clinical modalities rather than dominance by the largest test subset.

\subsection{Comparison Methods and Implementation}
We compare \ours{} with thirteen representative restoration methods: TAT \cite{Yang2025TATTaskAdaptiveTransformer}, Restormer \cite{Zamir2021RestormerEfficientTransformer}, SwinIR \cite{Liang2021SwinIRImageRestoration}, DaPT \cite{Wei2025DegradationAwarePromptedTransformer}, RAT \cite{Yang2024RegionAttentionTransformer}, PromptIR \cite{potlapalli2023promptir}, NAFNet \cite{chen2022simple}, AMIR \cite{Yang2024AllInOneMedicalImage}, DATPRL-IR \cite{dong2026datprlir}, ARGAN \cite{Luo2021Adaptiverectificationbased}, DenoMamba \cite{Ozturk2024DenoMambaFusedStateSpace}, MambaIR \cite{Guo2024MambaIRASimple}, and AirNet \cite{Li2022AllInOneImageRestoration}. All comparison models are re-trained and evaluated under the same PET/CT/MRI split. Table~\ref{tab:main_results} is therefore a reproduced-protocol comparison. Paper-reported or separately released official numbers are not mixed into this table because training schedules and checkpoint-selection rules may differ. For each method, Table~\ref{tab:main_results} reports the stronger checkpoint according to sample-weighted average PSNR.

For fair comparison, each method is evaluated with both the latest checkpoint and the best validation checkpoint when available, and the stronger checkpoint under sample-weighted average PSNR is reported. All metrics are computed on the full test set rather than a sampled subset. The evaluation script uses the same restored outputs and target images to compute PSNR, SSIM, and RMSE. Therefore, the metric differences in Tables~\ref{tab:main_results} and~\ref{tab:ablation} come from model outputs rather than from different post-processing pipelines.

\subsection{Implementation Details}
\ours{} is trained with uniform modality sampling, batch size 5 per modality, and effective batch size 15. Training uses $128\times128$ random crops, AdamW with learning rate $2\times10^{-4}$, betas $(0.9,0.99)$, weight decay 0, cosine annealing to $10^{-7}$, 1,000 warmup iterations, and gradient clipping at 1.0. The maximum schedule is 250k iterations. Validation is run every 1,000 iterations on up to 500 validation samples per modality, using all 128 CT validation slices because fewer than 500 are available. Early stopping is triggered after 20 validation rounds without improvement in the weighted validation score, with weights PET:1.0, CT:1.2, and MRI:1.5. The final tested \ours{} checkpoint is selected at 26,000 iterations.

The comparison methods are trained in independent environments with the same data paths and split files. This protocol is stricter than reporting only a final checkpoint because all-in-one restoration models may peak at different iterations for different modalities. Reporting the stronger checkpoint per method reduces the chance that a baseline is penalized by checkpoint timing.

\subsection{Main Results}
\begin{figure*}[t]
\centering
\includegraphics[width=0.95\textwidth]{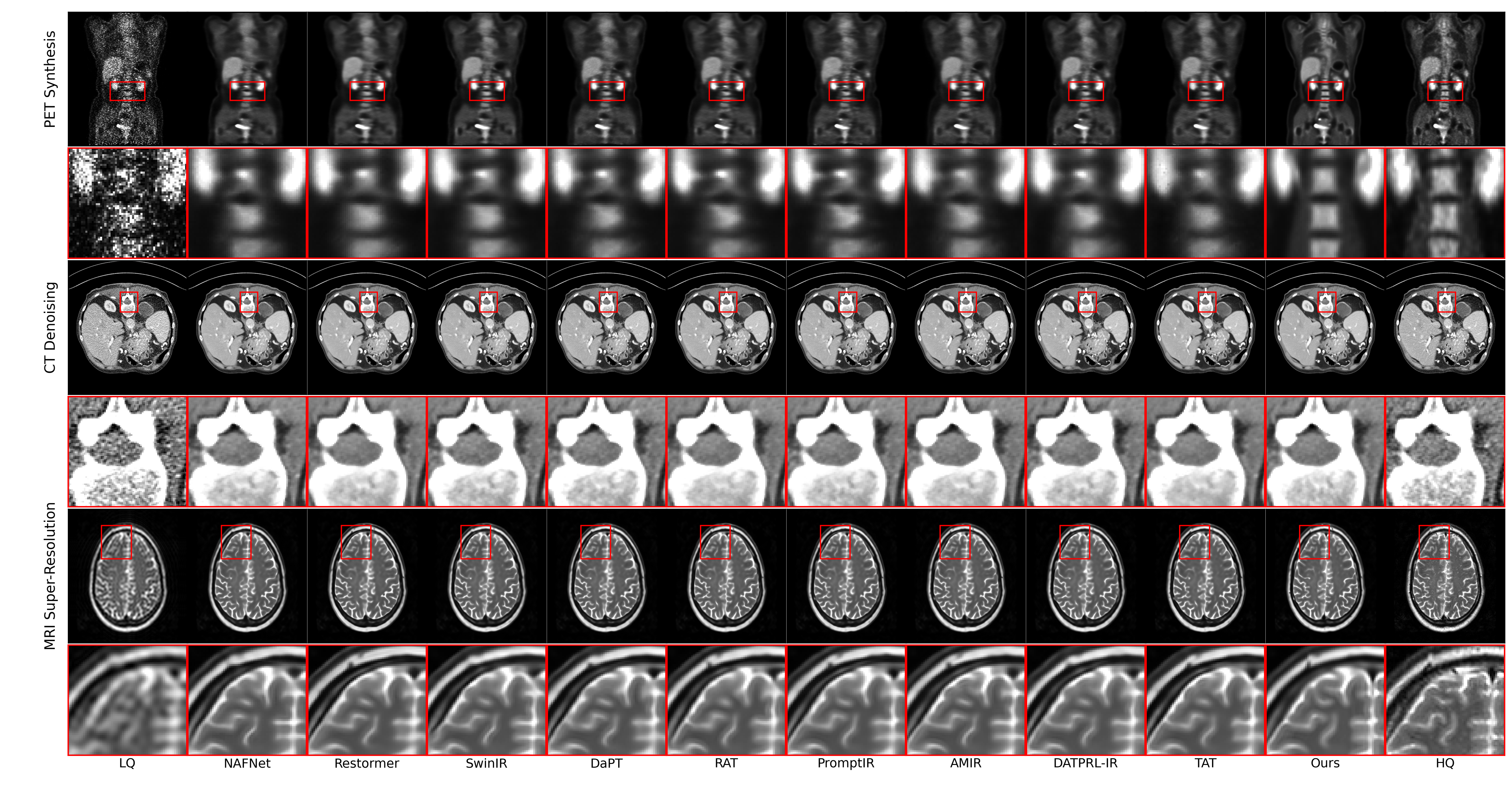}
\caption{Visual comparisons on all-in-one medical image restoration. Each row shows LQ input, representative comparison methods, \ours{}, and HQ target with matched zoom-in regions.}
\label{fig:visual_comparison}
\end{figure*}

Table~\ref{tab:main_results} reports PET synthesis, CT denoising, MRI super-resolution, and modality-balanced average results. \ours{} achieves the best table-level performance on all three modalities within this reproduced comparison. Compared with the strongest non-\ours{} PET result, \ours{} improves PET PSNR from 37.25 dB to 37.34 dB. On CT, where the reproduced baselines are tightly clustered, \ours{} improves PSNR from 33.79 dB to 33.85 dB. On MRI, \ours{} improves the best comparison result from 32.05 dB to 32.09 dB. The modality-average PSNR increases from 34.36 dB to 34.43 dB. The same table-level comparison shows consistent modality-average gains in SSIM and RMSE: average SSIM improves from 0.9357 to 0.9361, and average RMSE decreases from 12.4940 to 12.4811. Fig.~\ref{fig:visual_comparison} further provides qualitative comparisons across the three modalities.

\subsection{Mechanism Diagnostics}
\begin{figure*}[t]
\centering
\includegraphics[width=0.95\textwidth]{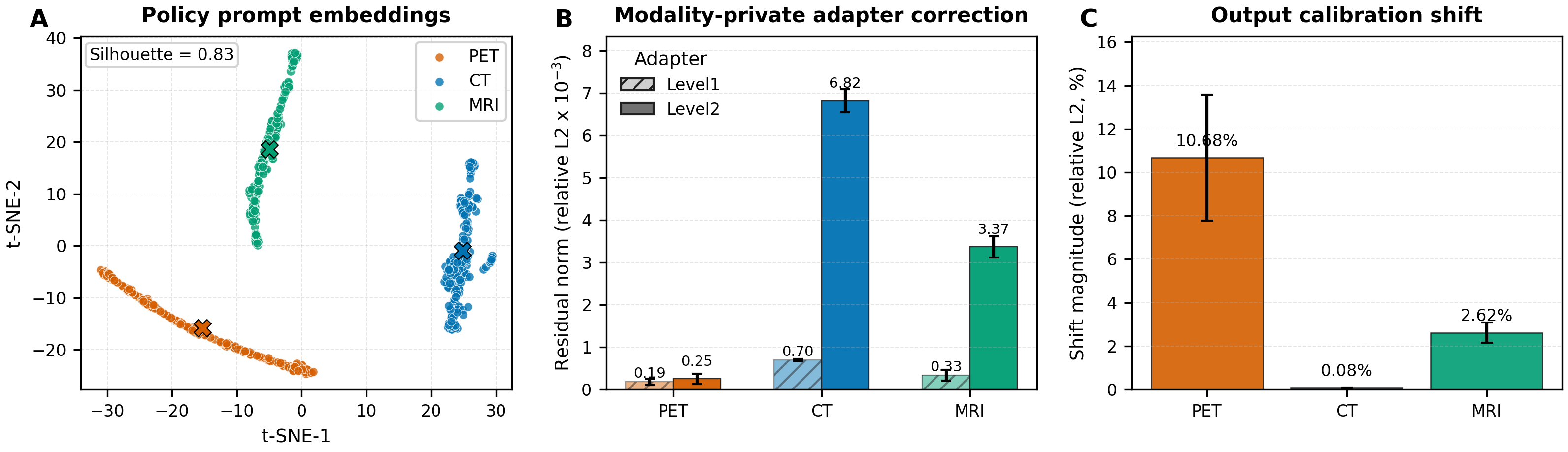}
\caption{Mechanism diagnostics of \ours{}. (A) Policy prompt embeddings visualized by t-SNE form modality-aware clusters. (B) Adapter residual norms show stronger level-2 feature correction for CT and MRI. (C) Output calibration shifts are largest for PET, indicating complementary image-domain residual correction.}
\label{fig:mechanism_diagnostics}
\end{figure*}

We further inspect the selected checkpoint without additional training by recording prompt embeddings, adapter residuals, and output-head residual shifts on 200 test samples per modality. Fig.~\ref{fig:mechanism_diagnostics} shows that the learned policy embeddings form clear modality-aware clusters in t-SNE space, with a silhouette score of 0.83. This pattern supports the use of the policy prompt as a modality-structured control representation. The adapter residuals are most active at decoder level 2 for CT and MRI, indicating that modality-private adapters mainly correct intermediate structural features for these modalities. In contrast, the output calibration shift is largest for PET, moderate for MRI, and minimal for CT, suggesting that PET benefits more from final image-domain residual calibration. These diagnostics are descriptive rather than causal, but they are consistent with the proposed decision-execution-correction decomposition.

\subsection{Component Analysis}
Table~\ref{tab:ablation} studies the contribution of output calibration, modality-private adaptation, and several extended designs. The goal of this ablation is not to show that every branch contributes equally, but to identify which residual error each constrained path absorbs. Output calibration alone already provides strong PET and CT results, indicating that image-domain residual correction is an important source of gain for modalities with visible intensity or noise bias. However, it is not a universal replacement for feature adaptation: the private-adapter variant improves the single-role MRI result over output calibration alone, consistent with MRI requiring structural feature refinement. Combining private adaptation and output calibration gives the best table-level results, including the best MRI score and modality-average PSNR, although the margin over the strongest single-role variants is small. Some extended variants increase MRI metrics relative to single-role variants, but none surpass the compact \ours{} design; their PET or CT trade-offs lower the balanced average. This pattern suggests that the value of \ours{} lies not in stacking all specialization modules, but in assigning limited adaptation capacity to complementary decision, feature, and output roles.

\subsection{Discussion}
The empirical results align with the proposed role-constrained design. At the decision level, Fig.~\ref{fig:mechanism_diagnostics}(A) shows that policy embeddings form modality-structured clusters, indicating that the prompt branch learns a useful control representation. At the feature-execution level, Fig.~\ref{fig:mechanism_diagnostics}(B) shows stronger level-2 adapter residuals for CT and MRI, consistent with their need for structural and anatomical refinement. At the output-calibration level, Fig.~\ref{fig:mechanism_diagnostics}(C) shows the largest calibration shift for PET, consistent with PET's low-count noise and image-domain intensity residuals. This pattern also explains the uneven gains in Table~\ref{tab:main_results}: PET and MRI benefit from larger correction margins, whereas CT baselines are already close and the remaining gain appears mainly as a small residual-bias reduction. Importantly, the ablation does not support a single dominant mechanism across modalities. Calibration explains much of the PET/CT improvement, while feature-level adaptation is more relevant to MRI. A compact decision-execution-calibration design is therefore a practical compromise for all-in-one deployment, because frequency and anchor variants can improve MRI while lowering PET or CT enough to reduce the balanced average.

\subsection{Limitations}
This study has several limitations. First, the current evaluation focuses on full-reference restoration metrics. PSNR, SSIM, and RMSE are appropriate for paired restoration benchmarks, but they do not directly measure downstream clinical utility. Future work should evaluate how restored images affect segmentation, quantification, and diagnosis. Second, \ours{} uses modality labels during training and testing. This assumption is reasonable for clinical imaging data, where modality metadata is normally available, but fully automatic modality inference could make the framework more robust to incomplete metadata. Finally, the proposed decomposition is supported by ablations and diagnostics rather than by a formal domain-shift bound. The current framework also uses lightweight statistic cues rather than a learned uncertainty estimator, and uncertainty-aware calibration may further improve reliability in difficult cases.

\section{Conclusion}
We presented \ours{}, a modality-aware residual recalibration framework for practical all-in-one medical image restoration. The method organizes unified PET, CT, and MRI restoration into degradation-aware prompt policy selection, prompt-gated modality-private residual adaptation, and zero-initialized output calibration. This role-constrained design provides a compact mechanism for balancing common restoration priors with modality-specific residual allocation in one model. Experiments on PET, CT, and MRI restoration show that \ours{} improves modality-balanced PSNR, SSIM, and RMSE over thirteen methods re-trained under the same protocol. The analysis suggests that the gains come from complementary adaptation roles rather than from a single uniformly dominant module. Future work will extend the framework to additional modalities and investigate uncertainty-aware calibration for clinically sensitive restoration errors.

\FloatBarrier
\bibliographystyle{IEEEtran}
\bibliography{references}

@inproceedings{Liang2021SwinIRImageRestoration,
  title = {SwinIR: Image Restoration Using Swin Transformer},
  author = {Liang, Jingyun and Cao, Jie and Sun, Guolei and Zhang, Kai and Van Gool, Luc and Timofte, Radu},
  year = {2021},
  booktitle = {Proceedings of the IEEE/CVF International Conference on Computer Vision Workshops},
  doi = {10.1109/ICCVW54120.2021.00210},
  eprint = {2108.10257}
}

@inproceedings{Zamir2021RestormerEfficientTransformer,
  title = {Restormer: Efficient Transformer for High-Resolution Image Restoration},
  author = {Zamir, Syed Waqas and Arora, Aditya and Khan, Salman and Hayat, Munawar and Khan, Fahad Shahbaz and Yang, Ming-Hsuan},
  year = {2022},
  booktitle = {Proceedings of the IEEE/CVF Conference on Computer Vision and Pattern Recognition},
  doi = {10.1109/CVPR52688.2022.00564},
  eprint = {2111.09881}
}

@inproceedings{chen2022simple,
  title = {Simple Baselines for Image Restoration},
  author = {Chen, Liangyu and Chu, Xiaojie and Zhang, Xiangyu and Sun, Jian},
  booktitle = {Computer Vision -- ECCV 2022},
  year = {2022},
  pages = {17--33},
  publisher = {Springer Nature Switzerland},
  doi = {10.1007/978-3-031-20071-7_2},
  eprint = {2204.04676}
}

@inproceedings{potlapalli2023promptir,
  title = {PromptIR: Prompting for All-in-One Image Restoration},
  author = {Potlapalli, Vaishnav and Zamir, Syed Waqas and Khan, Salman and Khan, Fahad Shahbaz},
  booktitle = {Advances in Neural Information Processing Systems 36},
  year = {2023},
  pages = {71275--71293},
  publisher = {Neural Information Processing Systems Foundation, Inc. (NeurIPS)},
  doi = {10.52202/075280-3121}
}

@inproceedings{Li2022AllInOneImageRestoration,
  title = {All-in-One Image Restoration for Unknown Corruption},
  author = {Li, Boyun and Liu, Xiao and Hu, Peng and Wu, Zhongqin and Lv, Jiancheng and Peng, Xi},
  booktitle = {2022 IEEE/CVF Conference on Computer Vision and Pattern Recognition (CVPR)},
  year = {2022},
  pages = {17431--17441},
  publisher = {IEEE},
  doi = {10.1109/CVPR52688.2022.01693}
}

@inproceedings{Yang2024AllInOneMedicalImage,
  title = {All-In-One Medical Image Restoration via Task-Adaptive Routing},
  author = {Yang, Zhiwen and Chen, Haowei and Qian, Ziniu and Yi, Yang and Zhang, Hui and Zhao, Dan and Wei, Bingzheng and Xu, Yan},
  year = {2024},
  booktitle = {Medical Image Computing and Computer Assisted Intervention -- MICCAI 2024},
  pages = {67--77},
  publisher = {Springer Nature Switzerland},
  doi = {10.1007/978-3-031-72104-5_7},
  eprint = {2405.19769}
}

@inproceedings{Yang2024RegionAttentionTransformer,
  title = {Region Attention Transformer for Medical Image Restoration},
  author = {Yang, Zhiwen and Chen, Haowei and Qian, Ziniu and Zhou, Yang and Zhang, Hui and Zhao, Dan and Wei, Bingzheng and Xu, Yan},
  year = {2024},
  booktitle = {Medical Image Computing and Computer Assisted Intervention -- MICCAI 2024},
  pages = {603--613},
  publisher = {Springer Nature Switzerland},
  doi = {10.1007/978-3-031-72104-5_58},
  eprint = {2407.09268}
}

@inproceedings{Yang2025TATTaskAdaptiveTransformer,
  title = {TAT: Task-Adaptive Transformer for All-in-One Medical Image Restoration},
  author = {Yang, Zhiwen and Zhang, Jiaju and Yi, Yang and Liang, Jian and Wei, Bingzheng and Xu, Yan},
  year = {2025},
  booktitle = {International Conference on Medical Image Computing and Computer-Assisted Intervention},
  pages = {565--575},
  publisher = {Springer Nature Switzerland},
  doi = {10.1007/978-3-032-05325-1_54},
  eprint = {2512.14550}
}

@article{Wei2025DegradationAwarePromptedTransformer,
  title = {Degradation-Aware Prompted Transformer for Unified Medical Image Restoration},
  author = {Wei, Jinbao and Yang, Gang and Wang, Zhijie and Tao, Shimin and Liu, Aiping and Chen, Xun},
  year = {2025},
  journal = {IEEE Transactions on Image Processing},
  doi = {10.1109/TIP.2025.3644795}
}

@inproceedings{Vaswani2017AttentionisAll,
  title = {Attention Is All You Need},
  author = {Vaswani, Ashish and Shazeer, Noam and Parmar, Niki and Uszkoreit, Jakob and Jones, Llion and Gomez, Aidan N. and Kaiser, Lukasz and Polosukhin, Illia},
  year = {2017},
  booktitle = {Advances in Neural Information Processing Systems},
  eprint = {1706.03762}
}

@inproceedings{Dosovitskiy2020AnImageis,
  title = {An Image Is Worth 16x16 Words: Transformers for Image Recognition at Scale},
  author = {Dosovitskiy, Alexey and Beyer, Lucas and Kolesnikov, Alexander and Weissenborn, Dirk and Zhai, Xiaohua and Unterthiner, Thomas and Dehghani, Mostafa and Minderer, Matthias and Heigold, Georg and Gelly, Sylvain and Uszkoreit, Jakob and Houlsby, Neil},
  year = {2021},
  booktitle = {International Conference on Learning Representations},
  eprint = {2010.11929}
}

@inproceedings{Liu2021SwinTransformerHierarchical,
  title = {Swin Transformer: Hierarchical Vision Transformer Using Shifted Windows},
  author = {Liu, Ze and Lin, Yutong and Cao, Yue and Hu, Han and Wei, Yixuan and Zhang, Zheng and Lin, Stephen and Guo, Baining},
  year = {2021},
  booktitle = {Proceedings of the IEEE/CVF International Conference on Computer Vision},
  doi = {10.1109/ICCV48922.2021.00986},
  eprint = {2103.14030}
}

@article{Gu2023MambaLinearTimeSequence,
  title = {Mamba: Linear-Time Sequence Modeling with Selective State Spaces},
  author = {Gu, Albert and Dao, Tri},
  year = {2023},
  journal = {arXiv preprint arXiv:2312.00752},
  eprint = {2312.00752}
}

@inproceedings{Zhu2024VisionMambaEfficient,
  title = {Vision Mamba: Efficient Visual Representation Learning with Bidirectional State Space Model},
  author = {Zhu, Lianghui and Liao, Bencheng and Zhang, Qian and Wang, Xinlong and Liu, Wenyu and Wang, Xinggang},
  year = {2024},
  booktitle = {International Conference on Machine Learning},
  eprint = {2401.09417}
}

@inproceedings{Liu2024VMambaVisualState,
  title = {VMamba: Visual State Space Model},
  author = {Liu, Yue and Tian, Yunjie and Zhao, Yuzhong and Yu, Hongtian and Xie, Lingxi and Wang, Yaowei and Ye, Qixiang and Jiao, Jianbin and Liu, Yunfan},
  year = {2024},
  booktitle = {Advances in Neural Information Processing Systems},
  pages = {103031--103063},
  publisher = {Neural Information Processing Systems Foundation, Inc. (NeurIPS)},
  doi = {10.52202/079017-3273},
  eprint = {2401.10166}
}

@inproceedings{Guo2024MambaIRASimple,
  title = {MambaIR: A Simple Baseline for Image Restoration with State-Space Model},
  author = {Guo, Hang and Li, Jinmin and Dai, Tao and Ouyang, Zhihao and Ren, Xudong and Xia, Shu-Tao},
  year = {2024},
  booktitle = {European Conference on Computer Vision},
  pages = {222--241},
  publisher = {Springer Nature Switzerland},
  doi = {10.1007/978-3-031-72649-1_13},
  eprint = {2402.15648}
}

@misc{Ozturk2024DenoMambaFusedStateSpace,
  title = {DenoMamba: A Fused State-Space Model for Low-Dose CT Denoising},
  author = {{\c{S}}aban {\"O}zt{\"u}rk and O{\u{g}}uz Can Duran and Tolga {\c{C}}ukur},
  year = {2024},
  eprint = {2409.13094},
  archivePrefix = {arXiv},
  primaryClass = {eess.IV},
  doi = {10.48550/arXiv.2409.13094},
  url = {https://arxiv.org/abs/2409.13094}
}

@inproceedings{Guo2024MambaIRv2AttentiveState,
  title = {MambaIRv2: Attentive State Space Restoration},
  author = {Guo, Hang and Guo, Yong and Zha, Yaohua and Zhang, Yulun and Li, Wenbo and Dai, Tao and Xia, Shu-Tao and Li, Yawei},
  year = {2025},
  booktitle = {Proceedings of the IEEE/CVF Conference on Computer Vision and Pattern Recognition},
  pages = {28124--28133},
  publisher = {IEEE},
  doi = {10.1109/CVPR52734.2025.02619},
  eprint = {2411.15269}
}

@article{Yang2024RestoreRWKVEfficientand,
  title = {Restore-RWKV: Efficient and Effective Medical Image Restoration with RWKV},
  author = {Yang, Zhiwen and Zhang, Hui and Zhao, Dan and Li, Jiayin and Wei, Bingzheng and Xu, Yan},
  year = {2026},
  journal = {IEEE Journal of Biomedical and Health Informatics},
  volume = {30},
  number = {1},
  pages = {513--526},
  publisher = {IEEE},
  doi = {10.1109/JBHI.2025.3588555},
  eprint = {2407.11087}
}

@inproceedings{Chan2018NoiseAdaptiveDeep,
  title = {Noise Adaptive Deep Convolutional Neural Network for Whole-Body PET Denoising},
  author = {Chan, Chung and Zhou, Jian and Yang, Li and Qi, Wenyuan and Kolthammer, Jeffrey and Asma, Evren},
  year = {2018},
  booktitle = {IEEE Nuclear Science Symposium and Medical Imaging Conference},
  doi = {10.1109/NSSMIC.2018.8824303}
}

@article{Luo2021Adaptiverectificationbased,
  title = {Adaptive Rectification Based Adversarial Network with Spectrum Constraint for High-Quality PET Image Synthesis},
  author = {Luo, Yanmei and Zhou, Luping and Zhan, Bo and Wang, Fei and Zhou, Jiliu and Wang, Yan and Shen, Dinggang},
  year = {2022},
  journal = {Medical Image Analysis},
  volume = {77},
  pages = {102335},
  publisher = {Elsevier},
  doi = {10.1016/j.media.2021.102335}
}

@article{Zhou20223DSegmentationGuided,
  title = {3D Segmentation Guided Style-Based Generative Adversarial Networks for PET Synthesis},
  author = {Zhou, Yang and Yang, Zhiwen and Zhang, Hui and Chang, Edward and Fan, Yubo and Xu, Yan},
  year = {2022},
  journal = {IEEE Transactions on Medical Imaging},
  doi = {10.1109/TMI.2022.3156614},
  eprint = {2205.08887}
}

@inproceedings{Yang2023DRMCAGeneralist,
  title = {DRMC: A Generalist Model with Dynamic Routing for Multi-Center PET Image Synthesis},
  author = {Yang, Zhiwen and Zhou, Yang and Zhang, Hui and Wei, Bingzheng and Fan, Yubo and Xu, Yan},
  year = {2023},
  booktitle = {Medical Image Computing and Computer Assisted Intervention -- MICCAI 2023},
  pages = {36--46},
  publisher = {Springer Nature Switzerland},
  doi = {10.1007/978-3-031-43898-1_4},
  eprint = {2307.05249}
}

@article{Jing2026MAPDiffMultiAnchorGuided,
  title = {MAP-Diff: Multi-Anchor Guided Diffusion for Progressive 3D Whole-Body Low-Dose PET Denoising},
  author = {Jing, Peiyuan and Cheng, Chun-Wun and Yang, Liutao and Zhang, Zhenxuan and Lima, Thiago V. and Strobel, Klaus and Leimgruber, Antoine and Aviles-Rivero, Angelica I. and Yang, Guang and Montoya-Zegarra, Javier A.},
  year = {2026},
  journal = {arXiv preprint arXiv:2603.02012},
  eprint = {2603.02012}
}

@inproceedings{Guo2024TransformerEncoderDecoderwith,
  title = {Transformer Encoder-Decoder with Depth-Wise Separable Convolution for Low-Dose CT Denoising},
  author = {Guo, Weizhen and Yuan, Huaqiang and Li, Yakang and Li, Jianfang},
  year = {2024},
  booktitle = {2024 5th International Conference on Computer Engineering and Application (ICCEA)},
  pages = {1693--1698},
  publisher = {IEEE},
  doi = {10.1109/ICCEA62105.2024.10604076}
}

@article{Wang2025DualDomainSelfConsistencyEnhancedDeep,
  title = {Dual-Domain Self-Consistency-Enhanced Deep Unfolding Network for Accelerated MRI Reconstruction},
  author = {Wang, Zhijie and Wei, Jinbao and Yang, Gang and Liu, Aiping and Wei, Wei and Qiu, Bo and Chen, Xun},
  year = {2025},
  journal = {Computer Methods and Programs in Biomedicine},
  doi = {10.1016/j.cmpb.2025.108995}
}

@article{Liu2026HDFNetHybriddomainfusionnetwork,
  title = {HDFNet: Hybrid-Domain Fusion Network for Medical Image Restoration},
  author = {Liu, Yuqi and Lin, Liqu and Wang, Shunzhou and Chen, Si and Zeng, Chao and Jiang, Nanfeng and Wang, Dahan},
  year = {2026},
  journal = {Expert Systems with Applications},
  doi = {10.1016/j.eswa.2026.131268}
}

@inproceedings{Chen2025AllinOneMedicalImage,
  title = {All-in-One Medical Image Restoration with Latent Diffusion-Enhanced Vector-Quantized Codebook Prior},
  author = {Chen, Haowei and Yang, Zhiwen and Hou, Haotian and Zhang, Hui and Wei, Bingzheng and Zhou, Gang and Xu, Yan},
  year = {2025},
  booktitle = {International Conference on Medical Image Computing and Computer-Assisted Intervention},
  pages = {67--77},
  publisher = {Springer Nature Switzerland},
  doi = {10.1007/978-3-032-05325-1_7},
  eprint = {2507.19874}
}

@article{Tang2026FastandSlow,
  title = {Fast and Slow Diffusion with Fuzzy Contextual Representation for Medical Image Translation},
  author = {Tang, Yu and Zheng, Jiahao and Ren, Fangmin and Huang, Dan and Zeng, Dan and Wu, Dong},
  year = {2026},
  journal = {IEEE Transactions on Emerging Topics in Computational Intelligence},
  doi = {10.1109/TETCI.2025.3642000}
}

@inproceedings{Wei2025RethinkingDiffusionBridge,
  title = {Rethinking Diffusion Bridge Model with Dual Alignments for Medical Image Synthesis},
  author = {Wei, Jinbao and Chen, Yuhang and Wang, Zhijie and Yang, Gang and Tao, Shimin and Gao, Jian and Liu, Aiping and Chen, Xun},
  year = {2025},
  booktitle = {ACM Multimedia},
  doi = {10.1145/3746027.3754923}
}

@article{Cui2025FocalModulationfor,
  title = {Focal Modulation for Image Restoration},
  author = {Cui, Yuning and Ren, Wenqi and Knoll, Alois},
  year = {2025},
  journal = {International Journal of Computer Vision},
  doi = {10.1007/s11263-025-02589-y}
}

@article{Cui2026CDIRLoRAInspiredAttention,
  title = {CDIR: LoRA-Inspired Attention for Efficient Composite Degradation Image Restoration},
  author = {Cui, Yuning and Ren, Wenqi and Shi, Boxin and Gan, Jianhou and Knoll, Alois},
  year = {2026},
  journal = {IEEE Transactions on Image Processing},
  doi = {10.1109/TIP.2026.3682023}
}

@article{Xiao2026QualiTeacherQualityConditionedPseudoLabeling,
  title = {QualiTeacher: Quality-Conditioned Pseudo-Labeling for Real-World Image Restoration},
  author = {Xiao, Fengyang and Feng, Jingjia and Hu, Peng and Zhang, Dingming and Xu, Lei and Qin, Guanyi and Li, Lu and He, Chunming and Farsiu, Sina},
  year = {2026},
  journal = {arXiv preprint arXiv:2603.08030},
  doi = {10.48550/arXiv.2603.08030},
  eprint = {2603.08030}
}

@inproceedings{dong2026datprlir,
  title = {Learning Domain-Aware Task Prompt Representations for Multi-Domain All-in-One Image Restoration},
  author = {Dong, Guanglu and Li, Chunlei and Ren, Chao and Hu, Jingliang and Shi, Yilei and Zhu, Xiao Xiang and Mou, Lichao},
  booktitle = {International Conference on Learning Representations},
  year = {2026}
}

\end{document}